\def\year{2017}\relax
\documentclass[letterpaper]{article}
\usepackage{aaai17}
\usepackage{times}
\usepackage{helvet}
\usepackage{courier}
\usepackage{ctable}
\usepackage{amsmath}
\usepackage{amssymb}
\newcommand{\eg}{{e.g.}}

\begin{document}
%
\title{Transfer Learning for Deep Learning on Graph-Structured Data}
\author{Jaekoo Lee, Hyunjae Kim, Jongsun Lee, Sungroh Yoon\\
Electrical and Computer Engineering\\
Seoul National University\\
Seoul 08826, Republic of Korea\\
sryoon@snu.ac.kr
}

\maketitle
\begin{abstract}
Graphs provide a powerful means for representing complex interactions between entities. Recently, new deep learning approaches have emerged for representing and modeling graph-structured data while the conventional deep learning methods, such as convolutional neural networks and recurrent neural networks, have mainly focused on the grid-structured inputs of image and audio. Leveraged by representation learning capabilities, deep learning-based techniques can detect structural characteristics of graphs, giving promising results for graph applications. In this paper, we attempt to advance deep learning for graph-structured data by incorporating another component: transfer learning. By transferring the intrinsic geometric information learned in the source domain, our approach can construct a model for a new but related task in the target domain without collecting new data and without training a new model from scratch. We thoroughly tested our approach with large-scale real-world text data and confirmed the effectiveness of the proposed transfer learning framework for deep learning on graphs. According to our experiments, transfer learning is most effective when the source and target domains bear a high level of structural similarity in their graph representations.
\end{abstract}

\section{Introduction}
Recently, many deep neural network models have been adopted successfully in various fields~\cite{lecun2015deep,schmidhuber2015deep}. In particular, convolutional neural networks (CNN)~\cite{krizhevsky2012imagenet} for image and video recognition and recurrent neural networks (RNN)~\cite{sutskever2014sequence} for speech and natural language processing (NLP) often deliver unprecedented levels of performance. Deep learning has also triggered advances in implementing human-level intelligence (\eg, in the game of Go~\cite{silver2016mastering}).

\begin{figure*}
	\centering
	\includegraphics[width=0.9\textwidth]{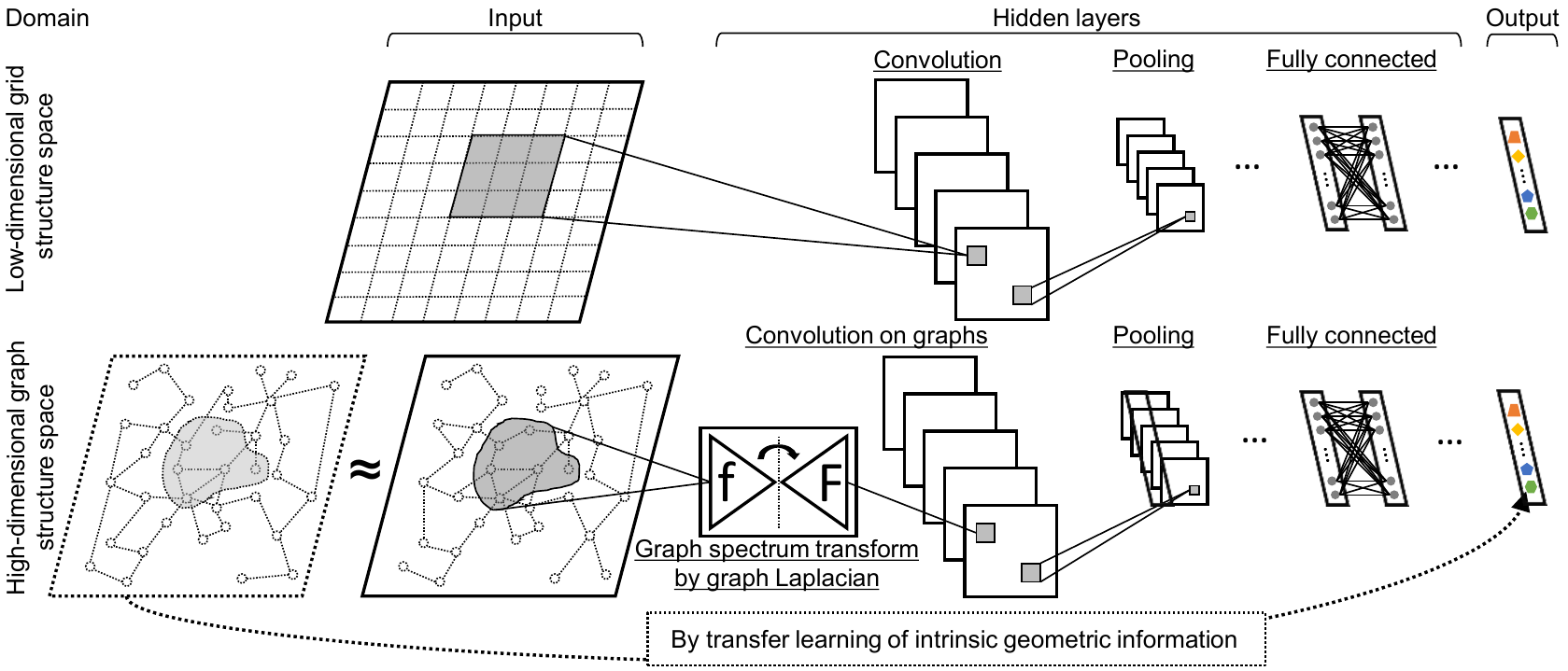}
	\caption{Conventional CNN works on a regular grid domain (top); proposed transfer learning framework for CNN, which can transfer intrinsic geometric information obtained from a source graph domain to a target graph domain (bottom).}
	\label{fig:comparison}
\end{figure*}

CNN and RNN extract data-driven features from input data (\eg, image, video, and audio data) structured in typically low-dimensional regular grids (see Fig.~\ref{fig:comparison}, top). Such grid structures are often assumed to have statistical characteristics (\eg, stationarity and locality) to facilitate the modeling process. Learning algorithms then take advantage of this assumption and boost performance by reducing the complexity of parameters~\cite{schmidhuber2015deep,bruna2013spectral,henaff2015deep}.

In reality, there exist a wide variety of data types in which we need more general non-grid structures to represent and model complex interactions among entities. Examples include social media mining and protein interaction studies. For such applications, a graph can provide a natural way of representing entities and their interactions~\cite{deo2016graph}. For graph-structured input, it is more challenging to find the statistical characteristics that can be assumed for grid-structured input~\cite{bruna2013spectral,henaff2015deep}.

Theoretical challenges including the above and practical limitations, such as data quantity/quality and training efficiency, make it difficult to apply conventional deep learning approaches directly, igniting research on adapting deep learning to graph-structured data~\cite{bruna2013spectral,henaff2015deep,jain2015structural,li2015gated}. In many graph analysis methods, the structural properties derived from input graphs play a crucial role in uncovering hidden patterns~\cite{koutra2013d,lee2015ricom}. The representation learning capability of deep networks is useful for automatically detecting data-driven structural features, and deep learning approaches have reported promising results.

In this paper, we attempt to advance deep learning for graph-structured data by incorporating another key component: transfer learning~\cite{pan2010survey}. By overcoming the common assumption that training and test data should be drawn from the same feature space and distribution, transfer learning between task domains can alleviate the burden of collecting data and training models for a new task. Given the importance of structural characteristics in graph analysis, the core of our proposal is to transfer the data-driven structural features learned by deep networks from a source domain to a target domain, as informally shown in Fig.~\ref{fig:comparison} (bottom). In the context of graphs, we call the transferred information the \emph{intrinsic geometric} information. 

Starting from this intuitive baseline, we need to fill in many details to implement transfer learning for deep learning on graph data. In particular, we need to answer two important questions: (\textbf{Q1}) under what condition can we expect a successful knowledge transfer between task domains and (\textbf{Q2}) how do we actually perform the transfer most effectively? This paper tries to address these questions.

To demonstrate the effectiveness of our approach, we tested it with large-scale public NLP datasets for text classification~\cite{zhang2015character}. Each dataset contained a corpus of news articles, Internet reviews, or ontology entries. We represented a dataset (\eg, Amazon reviews) with a graph to capture the interactions  among the words in the dataset. We then used the spectral CNN (SCNN)~\cite{bruna2013spectral,henaff2015deep} to model the graph using neural networks. The learned model can be used for classifying unseen texts from the same data source (Amazon). Furthermore, our experimental results confirmed that our transfer learning methodology allows us to implicitly derive a model for classifying texts from another source (\eg, Yelp reviews) without collecting new data and without repeating all the learning procedures from scratch.

Our specific contributions can be summarized as follows:
\begin{itemize}
	\item We proposed a new transfer learning framework for deep learning on input data in non-grid structure such as graphs. To the best of the authors' knowledge, this work is the first attempt of its kind. Adopting our approach will relieve the burden of re-collecting data and re-training models for related tasks.
	\item To address \textbf{Q1}, we investigated the conditions for successful knowledge transfers between graph domains. We conjectured that two graphs with similar structural characteristics would give better results and confirmed it by comparing graph similarity and transfer learning accuracy.
	\item To answer \textbf{Q2}, we tested diverse alternatives to the components of the proposed framework: graph generation, input representation, and deep network construction. In particular, to improve the SCNN model for extracting data-driven structural features from graphs, we analyzed and optimized the key factors that affect the performance of SCNN (\eg, the method to quantify spectral features of a graph).
	\item We performed an extensive set of experiments, using both synthetic and real-world data, to show the effectiveness of our approach. 
\end{itemize}

\section{Related Work}
Graphs can provide a general way of representing the diverse interactions of entities and have been studied extensively~\cite{sonawane2014graph}. In addition to studies on representation and quantification of relations and similarities~\cite{koutra2013d}, various studies focused on large-scale graph data and use of structural information. Recently, deep learning methods to automatically extract structural characteristics from graphs have been proposed~\cite{duvenaud2015convolutional,li2015gated}.

Examples of deep learning applied to non-grid, non-Euclidean space include graph wavelets from applying deep auto-encoders to graphs and using the properties of automatically extracted features~\cite{rustamov2013wavelets}, analysis of molecular fingerprints of proteins saved as graphs~\cite{duvenaud2015convolutional}, and a CNN-based model for handling tree structures in the context of programming language processing~\cite{mou2016convolutional}. 

Particularly relevant to our approach is the localized SCNN model~\cite{boscaini2015learning}, which is a deep learning approach that can extract the properties of deformable shapes. The generalized SCNN model~\cite{bruna2013spectral,henaff2015deep}, a key component of our framework, borrowed the Fourier transform concept from the signal processing field in order to apply CNNs in a grid domain to a graph-structured domain. In this model, the convolutional operation was re-defined for graphs. 

More recently, the restricted Boltzmann machine~\cite{lecun2015deep} was used to learn structural features from graphs in an unsupervised manner for classification~\cite{niepert2016learning}. For efficient and scalable semi-supervised classification of graph-structured data, the first-order approximation of spectral graph convolutions was utilized~\cite{kipf2016semi}. Our method could adopt these approaches as its base learning model to improve the effectiveness of transfer learning.

\section{Proposed Method}

\begin{figure*}
	\centering
	\includegraphics[width=0.86\textwidth]{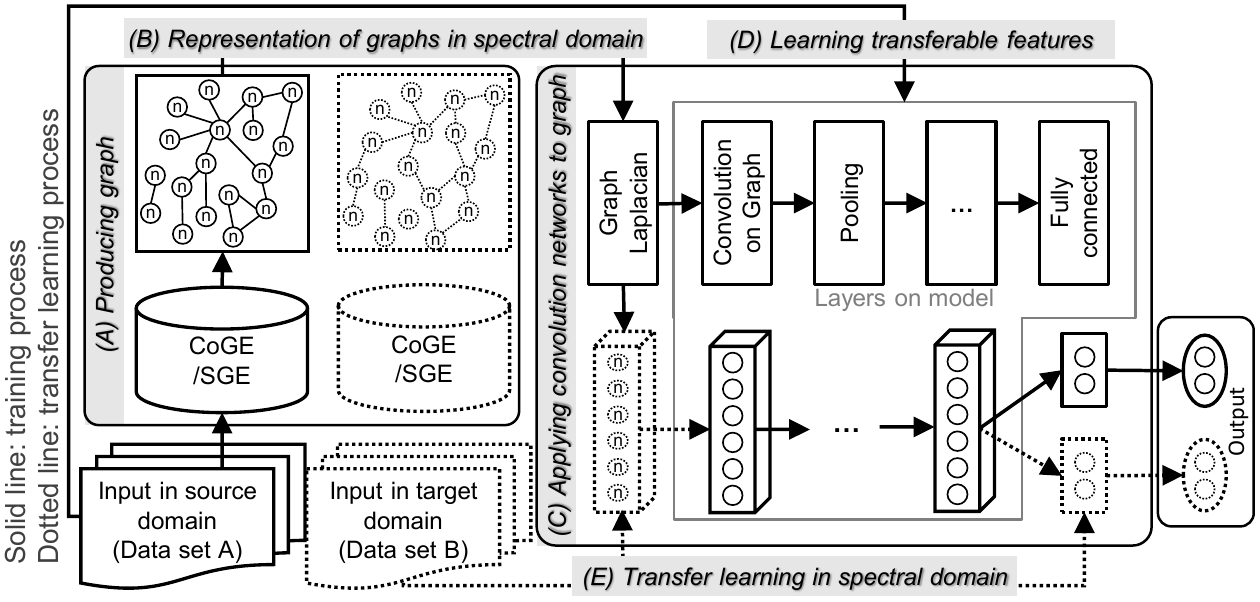}
	\caption{Overview of the proposed method.}
	\label{fig:overview}
\end{figure*}

Fig.~\ref{fig:overview} presents a diagram illustrating the overall flow of the proposed method, which consists of five steps, A--E. The first three steps are to produce a graph from input and to identify unique structural features from the graph. The last two steps are to apply transfer learning based on the learned features and graph similarity to carry out inference.

\subsection{Step A: Graph Production}
We represent data elements of input data and their interactions and relations as nodes and edges, respectively, in a graph. From an input dataset, we construct an undirected, connected, and weighted graph $G=(V,E,A)$, where $V$ and $E$ represent the sets of vertices and edges, respectively, and $A$ denotes the weighted adjacency matrix. Assume that $|V|=N$ and $|E|=M$.

We utilize two recent techniques to derive a graph (more specifically, the edge set $E$) from input data: co-occurrence graph estimation (CoGE)~\cite{sonawane2014graph} and supervised graph estimation (SGE)~\cite{henaff2015deep}. CoGE directly quantifies the closeness of data elements based on the frequency of co-occurrence, while SGE automatically learns a similarity features among elements through a fully connected network model.

\subsection{B: Representation of Graphs in Spectral Domain}

We extract the intrinsic geometric characteristics of the entire graph by deriving (non-)normalized Laplacian matrix $L$ of the graph constructed in step A. For a graph domain, $L$ provides the values for graph spectral bases in the convolution operation of SCNN~\cite{mohar1997some,koutra2013d}. 

We consider three types of $L$: the non-normalized Laplacian ($L^\text{basic}$), the random walk-based normalized Laplacian ($L^\text{rw}$), and the random walk with restart based normalized Laplacian ($L^\text{rwr}$) given by~\cite{tong2006fast}:
\begin{align}
L^\text{basic} & = D - A\\
L^\text{rw} & = D^{-1}(D-A)=I-D^{-1}A\\
L^\text{rwr} & = [I + \epsilon^{2}D - \epsilon A]^{-1} \approx [I- \epsilon A]^{-1}\\
&\approx I + \epsilon A + \epsilon^2 A^2 + \cdots
\end{align}
where $D$ represents the degree matrix\footnote{A diagonal matrix that shows the degree ({i.e.}, the number of edges attached) of each node.} of the graph, and $\epsilon$ represents the probability of restart. Note that the approximation in Eq. (3) is attained by attenuating neighboring influence, while the approximation in Eq. (4) is attained by belief propagation and its fast approximation.

$L$ is a symmetric matrix that can be decomposed through the diagonalization by combining eigenvalues $\lambda_{l}$ and the corresponding orthogonal eigenvectors $u_{l}(n)$, where $l$ is the order of an eigenvalue, and $n\in[1,N]$ is the index of a node~\cite{mohar1997some}.

Recall that a function $f:V \mapsto \mathbb{R}$ defined on the nodes of graph $G$ can be represented by a vector $f\in \mathbb{R}^N$ with the $n$-th dimension of $f$ indicating the  value at the $n$-th vertex in $V$~\cite{shuman2013emerging,shuman2016vertex}.
As in the Fourier transform, the eigenfunctions of $L$ represent the function $f$ defined by the nodes in the graph: $f_{G}(n)=\sum_{l=0}^{N-1}{\hat{f}_{G}(\lambda_l)u_l(n)} \leftrightarrow \hat{f}_{G}(\lambda_l)=\sum_{n=1}^{N}{f_{G}(n)u_l(n)}$, where $\hat{f}$, the transformed function of $f$, is represented by a set of basis eigenvectors. The Parseval's theorem also holds ({i.e.}, the energy of the transformed function is the same as that of the original function), and $\langle f, g \rangle=\langle \hat{f}, \hat{g} \rangle$ for two functions $f$ and $g$, verifying the consistency between the two domains~\cite{chung1997spectral,shuman2016vertex}.

This indicates that an input function defined on the vertex domain of a graph can be converted into the corresponding graph spectral domain by using the concept of Fourier analysis on graphs. The generalized convolutional operation (denoted by $\ast_{G}$) of functions $f$ and $g$ can be defined by the diagonalized linear multiplication in the spectral domain as follows~\cite{bruna2013spectral,henaff2015deep,shuman2016vertex}:
\begin{align*}
(f \ast_{G} g)(n)=\sum_{l=0}^{N-1}{\hat{f}(\lambda_l)\hat{g}(\lambda_l)u_l(n)}	
\end{align*}
which can also be expressed as
\begin{align}\label{eq:cov}
f \ast_{G} g = \hat{g}(L)f = U \left[\begin{array}{ccc}\hat{g}(\lambda_{0}) & \ldots & 0 \\
\vdots & \ddots & \vdots \\
0 & \ldots & \hat{g}(\lambda_{N-1})                                                                                                        \end{array} \right]U^{T}f	
\end{align}

\noindent where $U$ is a matrix having the eigenvectors of the graph Laplacian in its columns that quantify the intrinsic structural geometry of the entire graph domain and serve as the spectral bases of a graph. This matrix functions as the Fourier transform into the graph spectral domain. In this regard, a receptive filter learned through the convolution operation in a convolution layer of a CNN in a regular grid domain can be regarded as a matrix on $g$, which is diagonalized by $\hat{g}(\lambda_i)$ ($0 \le i \le N-1$) elements on input $f$ defined in the graph domain provided by Eq.~(\ref{eq:cov}).

For conventional CNNs, the $j$-th output $x_{k+1,j}$ from the $k$-th layer is defined as
\begin{align}
x_{k+1,j} = h\left(\sum_{i=1}^{\phi_{k-1}}{F_{k,i,j} \ast x_{k,i}}\right), \; j=1,\ldots,\phi_k
\end{align}
\noindent where $\phi_k$ is the number of feature maps, $x_{k}$ is the input of $k$-th layer, $h$ is a nonlinear function, and $F_{k,i,j}$ is the filter matrix from the $i$-th feature map to the $j$-th feature map.

For the SCNN, the transform of input $x_k$ of size $n \times \phi_{k-1}$ into output $x_{k+1}$ of size $n \times \phi_{k}$ is given by
\begin{align}
x_{k+1,j}=h\left(U\sum_{i=1}^{\phi_{k-1}}{F_{k,i,j}U^{T}x_{k,i}}\right), \; j=1,\ldots,\phi_k
\end{align}

\noindent where $h$ is a nonlinear function, and $F_{k,i,j}$ is a diagonal matrix. This implies that training the weights of learnable filters are the same as training the multipliers on the eigenvalues of the Laplacian~\cite{bruna2013spectral,henaff2015deep}. This characterizes the SCNN, a generalized CNN model that has several filter banks through generalized convolutional operations on a graph.

We augment the SCNN model so that it can support spatial locality, which is made independent of input size by using windowed smoothing filters. They are defined as $\hat{P}_{k}(l)=\sum_{k=0}^{K}{a_k \lambda_l^k}$ for $K<N$, based on the polynomial kernel $a_k$ with degree $K$~\cite{shuman2016vertex}. This is based on the fact (originally observed in signal processing) that the smoothness in the spectral domain can have spatial decay or local features in the original domain. We implement this idea using the eigenvectors of the subsampled Laplacian as the low-frequency eigenvectors of the Laplacian~\cite{boscaini2015learning}.

\subsection{C: Applying Convolutional Networks to Graphs}

We train the SCNN model by using the information obtained through the previous steps to represent the geometric information of local behaviors from the surface of a structural graph domain. The model has a hierarchical structure consisting of layers for convolutional and pooling, and a fully connected layer as shown in Fig.~\ref{fig:overview}. The training determines the weights of each layer by minimizing the task-specific cost (loss) function. The model can learn various data-driven features by re-defining the convolution operation with the spectral information of the structural graph domain~\cite{bruna2013spectral,henaff2015deep}. 

\subsection{D: Learning Transferable Features}
Once the model training is completed, it contains data-driven features for the graph-structured data derived from the input in steps A and B. As stated in Introduction, the core of our proposal is to transfer the information on structural characteristics of a graph learned by deep learning. The features learned in step C provide this information.

\subsection{E: Transfer Learning in Spectral Domain}
According to~\cite{pan2010survey}, a \textit{domain} in the context of transfer learning consists of a feature space $\mathcal{X}$ and a probability distribution $P(X)$, where $X \in \mathcal{X}$. Given a domain $\mathcal{D} =\{\mathcal{X}, P(X)\}$, we can denote a \textit{task} by $\mathcal{T}=\{\mathcal{Y}, f(\cdot) \}$ with a label space $\mathcal{Y}$ and a predictive function $f(\cdot)$  that is learned from training data $\{x,y\}$ where $x \in X$ and $y \in \mathcal{Y}$. The objective of general transfer learning is then to improve learning $f_T(\cdot)$ in the target domain $\mathcal{D}_T$ by exploiting the knowledge in the source domain $\mathcal{D}_S$ and task $\mathcal{T}_S$.

In the present context, we transfer the intrinsic geometric information learned from the graph $G_S$ encoding the knowledge in $\mathcal{D}_{S}$ and $\mathcal{T}_{S}$ in steps A--D. We skip the steps to generate $G_{T}$ for $\mathcal{T}_{T}$ in $\mathcal{D}_{T}$ as well as the steps to extract the structural characteristics therefrom. Under the condition that $G_S$ and $G_T$ bear structural similarities, we can directly build a model for $\mathcal{T}_{T}$ by (1) copying the convolutional and pooling layers that contain the features trained for $\mathcal{T}_{S}$ in $\mathcal{D}_{S}$, and by (2) training the fully connected layer of the model for fine tuning weights for $\mathcal{T}_{T}$ in $\mathcal{D}_T$.

This way of transfer learning provides efficiency in learning and also helps to minimize the problems resulting from lack of data and imperfect structural information for the new task. Note that the proposed method guarantees the spectral homogeneity of graphs by using the union of node sets on the heterogeneous source and target datasets. In our method, it is possible to utilize the spectral features of graphs from heterogeneous datasets.

\section{Results and Discussion}
We tested the proposed method by performing topic classification of text documents. Text data carry information on not only individual words but also on their relationships, and graph-based methods are widely used for text mining. We utilized large-scale public NLP data~\cite{zhang2015character}, which contained multiple corpora of news articles, Internet reviews, or ontology entries (Table~\ref{tab:datasets}). For controlled experiments, we also generated two pairs of synthetic datasets by random sampling of the real corpora. One pair consisted of two corpora with high similarity, and the other pair consisted of two corpora with low similarity.

\ctable[
caption     = Details of the real-world datasets used,
label       = tab:datasets,
doinside    = \footnotesize,
width       = \columnwidth,
]{lrrrr}{
	\tnote[]{\scriptsize {AG}: a corpus of news articles on the web; {DBP}: ontology data from DBpedia; YELP: reviews from Yelp; {AMAZ}: reviews from Amazon.}
	\tnote[$*$]{\scriptsize $\mathrm{sim}(G_1,G_2)=0$ indicates that two graphs $G_1$ and $G_2$ are structurally complementary, whereas the value of $1$ means that they are identical.}
	\tnote[$\dagger$]{\scriptsize $\mathrm{corr}(wv_1,wv_2)$ represents the correlation between the log-normalized bag of words extracted from each of the text corpora.}}{
	\toprule
	& AG & DBP & YELP & AMAZ \\
	\midrule
	Train & $120,000$ & $560,000$ & $580,000$ & $3,600,000$ \\
	Test  & $7,600$   & $70,000$  & $38,000$  & $400,000$ \\
	\#class & $4$       & $14$      & $2$       & $2$ \\
	\midrule
	Name & \multicolumn{4}{c}{$\mathrm{sim}(G_1,G_2)$\tmark[$^*$] [$\mathrm{corr}(wv_1,wv_2)$\tmark[$^\dagger$]]} \\
	\midrule
	AG &            & 0.37[0.45] & 0.28[0.36] & 0.35[0.42] \\
	DBP & 0.37[0.45] &             & 0.23[0.29] & 0.33[0.40] \\
	YELP & 0.28[0.36] & 0.23[0.29] &             & 0.50[0.58] \\
	AMAZ & 0.35[0.42] & 0.33[0.40] & 0.50[0.58] &            \\
	\bottomrule
}

\begin{figure*}
	\centering
	\includegraphics[width=\textwidth]{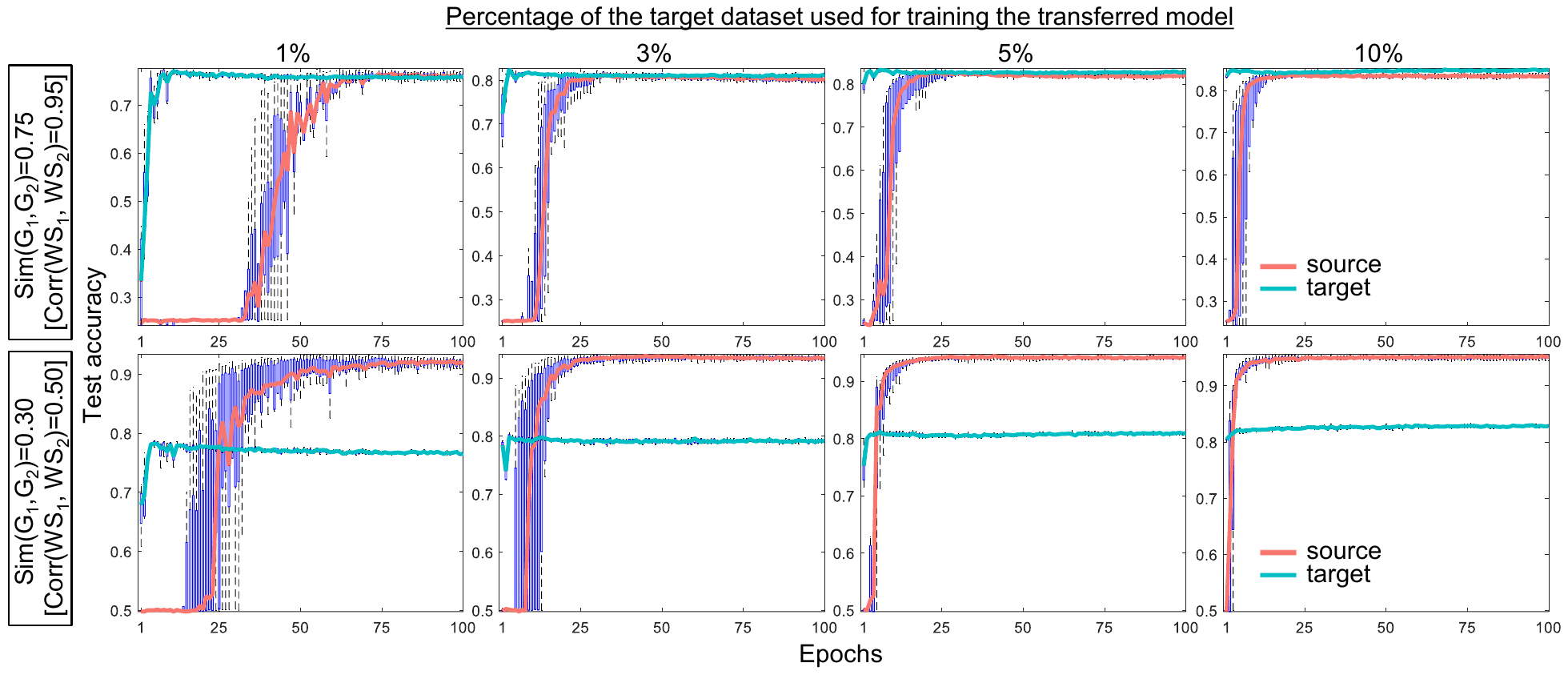}
	\caption{Results of intrinsic geometric information transfer learning for synthetic datasets (best viewed in color). Top: source and target datasets have high similarity in graph representations. Bottom: source and target datasets have low similarity. Each column: the percentage (1\%, 3\%, 5\% and 10\%) of the target dataset used for training the transferred model (fine tuning the fully connected layer). We repeated every experiment 10 times, and each data point shows a boxplot; red (source domain) and blue (target domain) lines connect the median locations of the boxplots.}
	\label{fig:ratio}
\end{figure*}

For measuring the structural similarity between graphs, as shown in Table~\ref{tab:datasets} and Fig.~\ref{fig:ratio}, we used the methods reported by existing studies~\cite{koutra2013d,lee2015ricom}; refer to the note below Table~\ref{tab:datasets} for more details. Note that YELP and AMAZ bear the highest similarity in terms of the metrics used.

We implemented the deep networks with Torch and Cuda using AdaGrad as the optimizer and ReLU as the activation. We carried out 10-fold cross validation. 
Note that the proposed method can offer an efficient training scheme with relatively low computation cost of $O(n^{2.376})$ by leaving out the eigenvalue decomposition on the SCNN and re-using the model trained by the data in the source domain. In our experiments, the proposed method provided more than 10\% reduction in the average training time.

Using the above setting, we first carried out comprehensive experiments to determine what factors affected the performance of the SCNN model for graph modeling. Table~\ref{tab:models} lists part of the results we obtained by varying the net architecture, the method to generate graphs, and the type of Laplacian matrix along with the resulting classification accuracy for each combination. We can observe from Table~\ref{tab:models} that the Laplacian methods do not significantly affect the performance, but $L^\text{rwr}$ had the benefit in terms of computational complexity. SGE tended to give more accurate results than CoGE, which implies that the initial graph generation affected the model training more critically than structural feature extraction. For the experiments shown in Table~\ref{tab:models}, the GC8-GC8-FC1K model (refer to the note below Table~\ref{tab:models} for notation) gave the best results, and we used this model as our main learning model in the following experiments.

\ctable[
caption     = Performance of SCNN model with various hyper-parameters for text topic classification task,
label       = tab:models,
doinside    = \scriptsize,
width       = \columnwidth,
]{lllcccc}{
	\tnote[$*$]{\scriptsize For training, we set the kernel degree $K = 60$, learning rate to 0.01 and used cross-entropy cost function with AdaGrad optimizer. GC8 means the use of graph convolutional layers with 8 feature maps, and FC500/FC1K means the use of fully connected layer with 500/1000 hidden units.}}{
	\toprule
	Model & Graph & $L$ & \multicolumn{4}{c}{Classification accuracy} \\
	architecture\tmark[$^*$]&gener.& type& {AG} & {DBP} & {YELP} & {AMAZ} \\
	\midrule
	GC8-FC500        & CoGE  & $L^\text{basic}$   & 0.89 & 0.95 & 0.91 & 0.88\\
	GC8-FC500        & CoGE  & $L^\text{rw}$      & 0.89 & 0.95 & 0.91 & 0.88\\
	GC8-FC500        & CoGE  & $L^\text{rwr}$     & 0.89 & 0.93 & 0.90 & 0.88\\
	GC8-FC500        & SGE   & $L^\text{basic}$   & 0.91 & 0.97 & 0.91 & 0.89\\
	GC8-FC500        & SGE   & $L^\text{rw}$      & 0.91 & 0.97 & 0.91 & 0.88\\
	GC8-FC500        & SGE   & $L^\text{rwr}$     & 0.89 & 0.95 & 0.91 & 0.89\\
	GC8-FC1K        & CoGE  & $L^\text{basic}$  & 0.90 & 0.95 & 0.93 & 0.88\\
	GC8-FC1K        & CoGE  & $L^\text{rw}$     & 0.90 &	0.97 & 0.92 & 0.89\\
	GC8-FC1K        & CoGE  & $L^\text{rwr}$    & 0.89 &	0.96 &	0.92 & 0.89\\
	GC8-FC1K        & SGE   & $L^\text{basic}$  & 0.91 &	0.97 &	0.92 & 0.88\\
	GC8-FC1K        & SGE   & $L^\text{rw}$     & 0.91 &	0.97 & 0.92 & 0.88\\
	GC8-FC1K        & SGE   & $L^\text{rwr}$    & 0.89 & 0.96 &	0.91 & 0.88\\
	GC8-GC8-FC1K    & CoGE  & $L^\text{basic}$  & 0.89 &	0.96 &	0.92 & 0.89\\
	GC8-GC8-FC1K    & CoGE  & $L^\text{rw}$     & 0.89 & 0.97 &	0.92 & 0.89\\
	GC8-GC8-FC1K    & CoGE  & $L^\text{rwr}$    & 0.89 &	0.97 &	0.92 & 0.88\\
	GC8-GC8-FC1K    & SGE   & $L^\text{basic}$  & 0.91 &	0.97 &	0.92 & 0.88\\
	GC8-GC8-FC1K    & SGE   & $L^\text{rw}$     & 0.91 &	0.97 &	0.92 & 0.88\\
	GC8-GC8-FC1K    & SGE   & $L^\text{rwr}$    & 0.89 &	0.97 &	0.92 & 0.89\\
	\bottomrule
}


We then performed experiments to determine the effectiveness of transfer learning using the synthetic datasets. The results are shown in Fig.~\ref{fig:ratio}; the plots in the top row are from the pair of synthetic corpora with high similarity [$\mathrm{sim}(G_1,G_2)=0.75$ and $\mathrm{corr}(wv_1,wv_2)=0.95$] for varying quantities of fine-tuning data for training the transferred model in the target domain (1\%, 3\%, 5\%, and 10\% of the entire target data). The plots in the bottom row of Fig.~\ref{fig:ratio} correspond to the results from the pair of synthetic corpora with low similarity [$\mathrm{sim}(G_1,G_2)=0.30$ and $\mathrm{corr}(wv_1,wv_2)=0.50$]. We can observe that transfer learning is more effective for the higher similarity case, in which the test accuracy of the transferred model increased significantly faster than that of the source domain model. Using only 1\% of the target domain data was sufficient for training, and using more data did not provide a noticeable difference. For the lower similarity case, the training in the target domain was limited and could not deliver the same level of accuracy in the source domain due to discrepancies in the underlying structure between the source and target domains.

\begin{figure*}
	\centering
	\includegraphics[width=\textwidth]{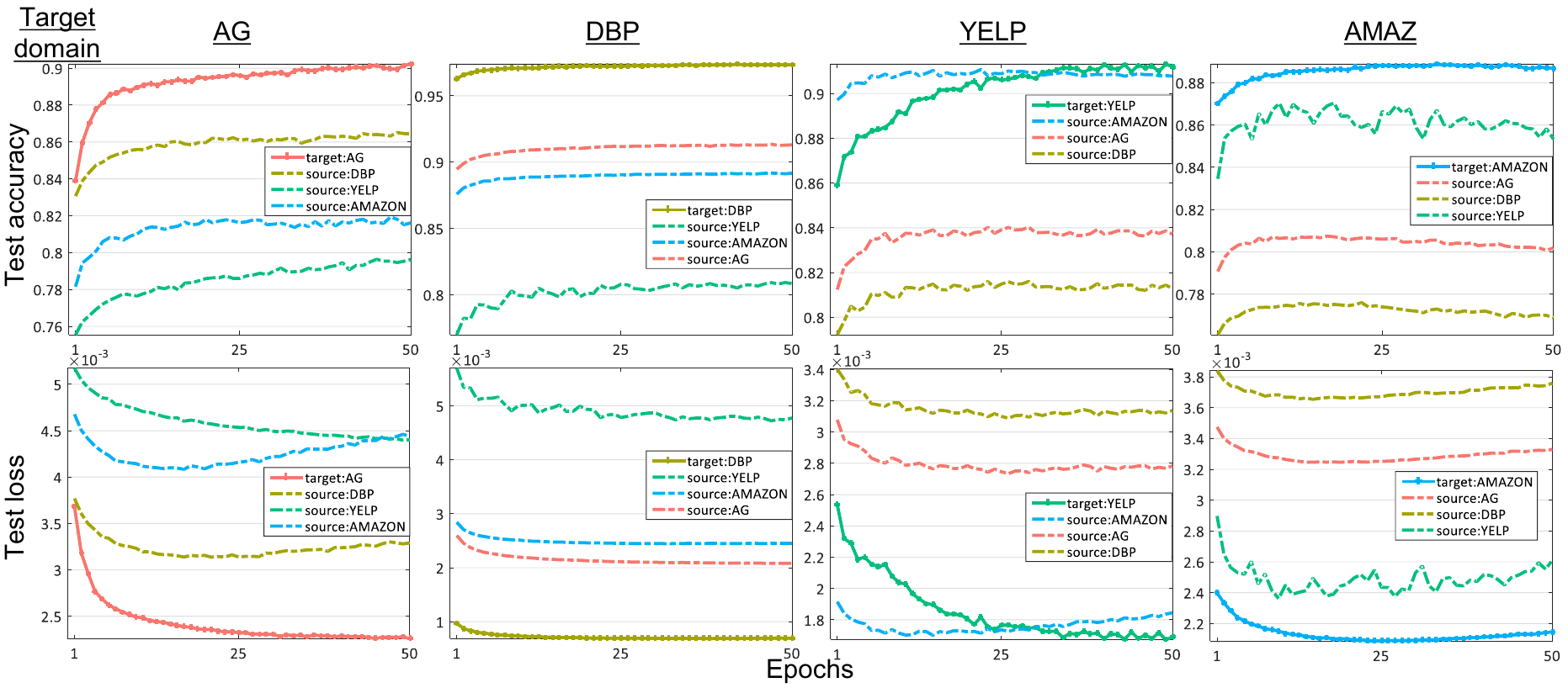}
	\caption{Results of the proposed method on real-world datasets (best viewed in color). Each plot in the top row: test accuracy of the model trained with the original data (solid lines) and those of the models trained with the other data sources and transferred (dotted lines). Each plot in the bottom row: test loss. For YELP and AMAZ, transfer learning was most effective, given that they have the highest level of structural similarity of all the cases (see Table~\ref{tab:datasets}).}
	\label{fig:datasets}
\end{figure*}

Finally, we tested our approach with four corpora (AG, DBP, YELP, and AMAZ) as shown in Fig.~\ref{fig:datasets}. The plots in the top row represent the test accuracy of the model trained with the original data (solid line) and those of the transferred model trained with each of the other data (dotted line). The bottom plots represent the test loss.
For the two corpora with the highest level of similarity (YELP and AMAZ), the effect of transfer learning was most salient. The test accuracy of the transferred model was comparable to that of the source model (for YELP) or was only 5--8\% lower (for AMAZ). For the other cases with lower similarity than YELP and AMAZ, transfer learning was less effective. These results again confirmed our observation that the knowledge transfer is most successful when the source and target domains have high level of structural similarity between underlying graph representations.

\section{Conclusion}
We have proposed a new transfer learning framework for deep learning on graph-structured data. Our approach can transfer the intrinsic geometric information learned from the graph representation of the source domain to the target domain. We observed that the knowledge transfer between tasks domains is most effective when the source and target domains possess high similarity in their graph representations. We anticipate that adoption of our methodology will help extend the territory of deep learning to data in non-grid structure as well as to cases with limited quantity and quality of data. To prove this, we are planning to apply our approach to diverse datasets in different domains.

\section{Acknowledgments}
This work was supported by Ministry of Science, ICT \& Future Planning [No.2016M3A7B4911115, No.PA-C000001], the Institute for Information Communications Technology Promotion [No.B0101-16-0307], the Brain Korea 21 Plus Project in 2016 grant funded by the Korea government, and Samsung Research Funding Center of Samsung Electronics [No.SRFC-IT1601-05].


\bibliographystyle{aaai}
\bibliography{references}

\end{document}